\documentclass{article}
\usepackage[utf8]{inputenc}
\usepackage{titling}
\usepackage{graphicx}
\usepackage{placeins}
\usepackage[left=1in,right=1in,top=1in,bottom=1in]{geometry}
\usepackage{amsmath}
\usepackage{amssymb}
\usepackage{xcolor}
\usepackage{url}
\usepackage{hyperref}
\usepackage{booktabs}
\usepackage[normalem]{ulem}
\usepackage{titling}
\usepackage{float}
\usepackage{subcaption}
\usepackage{comment}
\usepackage{authblk}

\title{Unsupervised Framework for Evaluating and Explaining Structural Node Embeddings of Graphs}

\author[1]{Ashkan Dehghan\thanks{ashkan.dehghan@torontomu.ca}}
\author[1,2]{Kinga Siuta}
\author[1,2]{Agata Skorupka}
\author[3]{Andrei Betlen}
\author[3]{David Miller}
\author[2]{Bogumi\l{} Kami\'nski\thanks{bkamins@sgh.waw.pl}}
\author[1]{Pawe\l{} Pra\l{}at\thanks{e-mail: pralat@torontomu.ca; part of this work was done while the author was visiting the Simons Institute for the Theory of Computing.}}
\affil[1]{Toronto Metropolitan University, Toronto, ON, Canada}
\affil[2]{SGH Warsaw School of Economics, Warsaw, Poland}
\affil[3]{Patagona Technologies, Pickering, ON, Canada}

\date{}

\begin{document}

\maketitle

\begin{abstract}
An embedding is a mapping from a set of nodes of a network into a real vector space. Embeddings can have various aims like capturing the underlying graph topology and structure, node-to-node relationship, or other relevant information about the graph, its subgraphs or nodes themselves. A practical challenge with using embeddings is that there are many available variants to choose from. Selecting a small set of most promising embeddings from the long list of possible options for a given task is challenging and often requires domain expertise. Embeddings can be categorized into two main types: classical embeddings and structural embeddings. Classical embeddings focus on learning both local and global proximity of nodes, while structural embeddings learn information specifically about the local structure of nodes' neighbourhood. For classical node embeddings there exists a framework which helps data scientists to identify (in an unsupervised way) a few embeddings that are worth further investigation. Unfortunately, no such framework exists for structural embeddings. In this paper we propose a framework for unsupervised ranking of structural graph embeddings. The proposed framework, apart from assigning an aggregate quality score for a structural embedding, additionally gives a data scientist insights into properties of this embedding. It produces information which predefined node features the embedding learns, how well it learns them, and which dimensions in the embedded space represent the predefined node features. Using this information the user gets a level of explainability to an otherwise complex black-box embedding algorithm.
\end{abstract}

\section{Introduction}\label{sec:intro}

Inspired by early work in word embedding techniques~\cite{mikolov2013efficient}, node, edge and graph embedding algorithms have gained a lot attention in the machine learning community, in recent years. Indeed, learning an accurate and useful latent representation from network-data is an important and necessary step for any successful machine learning task, including node classification~\cite{neville2000iterative}, anomaly detection~\cite{akoglu2015graph}, link prediction~\cite{hasan2011survey}, and community detection~\cite{pankratz_preprint} (see also surveys~\cite{chami2020machine, hamilton2017representation}). 

\medskip

In this paper, we distinguish two families of node embeddings: classical node embeddings and structural node embeddings. The first family is very rich with already over 100 algorithms proposed in the literature. Informally speaking, \emph{classical node embeddings} fall into broad and diverse family of embeddings that try to assign vectors in some high dimensional space to nodes of the graph that would allow for its approximate reconstruction using such encapsulated information. Different classical embedding algorithms use different approaches to achieve this task. Some of them, in order to extract useful information from graphs, try to create an embedding in a geometric space by assigning coordinates to each node such that nearby nodes are more likely to share an edge than those far from each other. Some other approaches postulate that pairs of nodes that have overlapping neighbourhoods (not necessarily intermediate ones) should have similar representations in the embedded space. Independently, the techniques to construct the desired classical embeddings can be broadly divided into the following three groups: linear algebra algorithms, random walk based algorithms, and deep learning methods~\cite{aggarwal2020machine,kaminski2021mining}. 

Classical embeddings work well for machine learning tasks such as link prediction. However, as the study of~\cite{stolman2022classic} shows, they do not guarantee good performance, for example, in tasks such as community labeling that can be viewed as a classification task or role detection. The reason is that in these challenging-for-classical-embeddings machine learning problems, when doing inference, it is important to preserve structural characteristics of nodes. Informally speaking again, by structural characteristics of nodes we mean the structure of nodes' egonets (that is, induced subgraphs of given nodes and their neighbourhoods up to some fixed depth) but disregarding node labelling. The simplest form of one-dimensional structural embeddings are node features such as degree or local clustering coefficient. Indeed, node features have been used extensively since the very beginning of network analysis, as most of them have natural interpretations and are usually relatively easy to compute. From the standpoint of this discussion, it is important to highlight that such node features do not depend on node labels, but rather on the relationships between them. For example, two nodes might both have large and comparable degrees or similar pageranks (and, as a result, end up close to each other in the embedded space) but be distant from each other in terms of concrete neighbours (and so they would be far apart in classical embeddings). The already mentioned study~\cite{stolman2022classic} shows that such node features are efficient in various tasks such as community labeling. The reason is that often the role of a node within a graph is an important predictor of some features but not necessarily its concrete neighbours.

Since using hand-crafted node features in various machine learning tools has proven to be a useful technique, a number of \emph{structural embedding algorithms} such as \textbf{RolX}~\cite{henderson2012rolx}, \textbf{Struct2Vec}~\cite{struc2vec}, \textbf{GraphWave}~\cite{donnat2018learning}, and \textbf{Role2Vec}~\cite{role2vec} have been recently proposed. This family of embeddings is much smaller but it is expected to grow in the near future. Such embedding algorithms try to capture structural characteristics of nodes, that is, put nodes that have similar structural characteristics close together in the embedded space. Again, like with classical embeddings, concrete implementations of structural embeddings differ in the way how they define similarity between neighbourhoods of two nodes (possibly of different sizes but similar to each other)---see Subsection~\ref{sec:motivation_embeddings} for a longer discussion.

\medskip

Most embedding algorithms, both classical and structural, have plenty of parameters to tune, the dimension of the embedding being only one of them but an important one. Moreover, most of them are randomized algorithms which means that even for a given graph, each time we run them we get a different embedding, possibly of different quality. Because of that, it is not clear which algorithm and parameters should be used for a given application at hand---even for the very same application, the decision of which algorithm to use might depend on the properties of the investigated network~\cite{dehghan2022evaluating}. 

In order to help data scientists to filter (in an unsupervised way) a large family of classical embeddings to choose from and identify a few embeddings that are worth further investigation, a framework was recently introduced~\cite{kaminski2021multi, kaminski2020unsupervised}. In order to handle enormous graphs, a landmark-based version of the framework was proposed in~\cite{kaminski2020scalable}, which can be calibrated to provide a good trade-off between performance and accuracy. The code is available on GitHub repository\footnote{\url{https://github.com/KrainskiL/CGE.jl}}. The framework assigns to each embedding two scores, local and global, that measure the quality of an evaluated embedding for tasks that require good representation of local and, respectively, global properties of the network. Unfortunately, there is no framework to evaluate structural embeddings (in an unsupervised way) and in this paper we propose one such approach.

\medskip

Suppose that a data scientist faces a problem in which a good structural embedding is required. As already commented above, there are many structural embeddings available to chose from, and each of them often has many parameters that can be tuned. Moreover, as will be discussed in Section~\ref{sec:motivation}, in general there is no uniformly best structural embedding, even for a single type of problem. Their effectiveness changes depending on various properties of the graph that needs to be dealt with. For these reasons, there is a need for an unsupervised way to distinguish good structural embedding from bad ones.

In general, structural embeddings are designed to capture many complex structural graph characteristics. However, in this work we stipulate that, as a minimum requirement, a good structural embedding should be able to reconstruct at least some selected node features that are commonly used in network analysis to capture their structural properties (such as node degree, local clustering coefficient, etc.). The assumption here is that data scientist using the framework is usually able to identify which standard node features might be useful to be preserved for a given machine learning task at hand. However, it is important to note that structural embeddings might capture more structural characteristics of the graph than the ones selected by a user. For this reason, we propose a framework that is able, in an unsupervised way, not only to create a ranking of embeddings explaining how well they allow for reconstruction of selected node features, but also identifies which dimensions of the embeddings are actually important for this reconstruction. This, implicitly, identifies other dimensions (that are not important to explain the selected node features) that carry information about structural characteristics of a graph that are not captured by the selected features.

\medskip

This paper is an extended version of the proceedings paper~\cite{dehghan1unsupervised}.
This paper is structured as follow:
\begin{itemize}
    \item In chapter 2, we use two real-world networks to investigate the differences between classical and structural node embeddings. We introduce a set of structural node features and show how classical node embeddings fail to capture such features.
    \item In chapter 3, we introduce our framework and outline the algorithm and its parameters.
    \item Finally in chapter 4, we use both a synthetic and a real network to showcase and analyze various algorithmic properties of our framework. We also highlight an example use case of our framework by exploring role classification case study.
\end{itemize}

\section{Motivation}\label{sec:motivation}
In order to motivate the line of research presented in this paper, we need to do two things. First of all, since there already exists a framework that evaluates classical node embeddings~\cite{kaminski2021multi}, we need to argue that structural graph embeddings have different properties than classical ones and so require a new evaluation process. Structural embeddings focus on capturing different properties of the network---see Subsection~\ref{sec:motivation_embeddings}. As a result, the two scores returned by the existing framework~\cite{kaminski2021multi} (despite the fact that they are useful in deciding which embeddings should be used in applications such as link prediction, clustering, and classification~\cite{dehghan2022evaluating}) are meaningless when it comes to applications that require structural information about graph nodes.  Once the reader is convinced that there is a need for another evaluation framework, the second task is to motivate the specific approach we took to design it---see Subsection~\ref{sec:motivation_framework}.

\subsection{Background on Structural Embedding Algorithms}\label{sec:motivation_embeddings}

As it was discussed in Section~\ref{sec:intro}, there are the two types of embeddings, classical and structural. They differ significantly when it comes to what they aim to capture. However, since the recent development of structural embeddings started independently by various groups of researchers, there is no agreement yet on what exactly they are supposed to capture. In this subsection, we briefly summarize different approaches present in the literature. 

In~\cite{donnat2018learning}, structural node discovery of a node is defined as a process that tries to identify nodes which have topologically similar network neighbourhoods while residing in potentially distant parts of the network. It is contrasted with the classical node embedding techniques, which try to place nodes that are close in the network to be also close in the embedded space. Similar view is shared in~\cite{struc2vec} where it is assumed that under structural embedding two nodes that have identical local network structures should have the same representation in the embedded space, while nodes with different structural identities should be far apart. In particular, the latent representation should \emph{not} depend on any node or edge attributes such as their labels. A related approach is used in~\cite{rossi2020}, where structural node embedding is defined using subgraph patterns present in a nodes' neighbourhoods such as network motifs, graphlet orbits/positions. In~\cite{role2vec}, the authors emphasized that classical embedding algorithms based on random walks on graphs mainly capture node proximity, and proposed a notion of a ``role'' for measuring structural similarity of nodes. Based on this notion, the authors introduced \textbf{Role2Vec}, a random walk based algorithm, which does not use node identity but rather considers nodes attributes instead.

However, other authors define the notion of structural similarity slightly differently. For example, in~\cite{jin2014} two nodes are defined to be structurally equivalent if the nodes have the same sets of neighbours. In this paper, we classify embeddings that would follow this notion of equivalence, as classical. However, in~\cite{jin2014} different notions of node equivalence are introduced that lead to structural embeddings according to the definition we use in this paper. First, they define automorphic equivalence of two nodes if there exists a graph automorphism that maps one of the nodes to the other (informally, two nodes can swap labels, along with possibly some label swaps of other nodes, while preserving all of the relationships in the graph). A more flexible notion of equivalence discussed in that work, that also leads to structural embeddings, is \emph{regular equivalence}, which concentrates only on having neighbours of some type, ignoring the number of such neighbours. As a result, two nodes of different degrees can be regularly equivalent if they have the same types of neighbours. Similar definitions are used, for example, in~\cite{rossi2015}.
A similar notion of structural similarity as proposed in~\cite{jin2014} is considered in~\cite{wang2016}. In~\cite{wang2016}, embeddings are said to be structure-preserving if they preserve second-order proximity of nodes, that is, two nodes are similar if their neighbours are similar in the sense of classical embeddings. Again, in this paper we classify such embeddings as classical. Finally, let us mention about one more paper that is closely related to the discussion we have in this subsection. In~\cite{rossi2020a}, the authors argue that graph embeddings should be divided into two classes: community-based or role-based. They discuss various properties and potential applications of both approaches. In this paper we take a similar approach by calling community embeddings \emph{classical} (as this class was developed earlier and received more attention from the researchers), and role-based embedding algorithms as \emph{structural}.

\subsection{Properties of Structural Embeddings}\label{sec:motivation_framework}

The previous subsection shows that there are various approaches used in the existing structural node embeddings, and more embeddings from this family will likely be introduced in the near future. Details of their design may vary but, in general, structural embeddings are designed to capture various complex structural graph characteristics. We believe that a good structural embedding should at least be able to reconstruct some selected node features that are commonly used in network analysis to capture their structural properties (such as node degree, local clustering coefficient, etc.). On the other hand, since the aim of classical embeddings is different, they might not be able to do that. In order to test this hypothesis and illustrate the difference between the two families of embeddings, we performed an experiment in which different embeddings of two selected real-life networks are analyzed to see if they are able to explain some commonly used node features.

\medskip

The experiments were performed on two datasets representing Bitcoin transactions networks on the two platforms (markets): OTC and Alpha. In the paper we will refer to them as \texttt{Bitcoin OTC}\footnote{\url{https://snap.stanford.edu/data/soc-sign-bitcoin-otc.html}} and \texttt{Bitcoin Alpha}\footnote{\url{https://snap.stanford.edu/data/soc-sign-bitcoin-alpha.html}}, respectively. In both networks, nodes represent users and edges represent the fact of being rated by a given user~\cite{kumar2018rev2}, thus the considered graphs are directed. Rating can happen when there is a transaction between users and the value (represented as edge weight) can achieve scores from $-10$ (total distrust) to $10$ (total trust), basing of how well the user performed during the transaction. User can be rated only once but the rating can be updated. Only users with positive ratings can rate other market players.

\texttt{Bitcoin OTC} is a directed graph with 5,881 nodes and 35,592 edges, whereas for \texttt{Bitcoin Alpha} these values are 3,783 and, respectively, 24,186. Some additional basic characteristics of both graph datasets are presented in Table~\ref{table:graph_stats}.

The main advantage of these datasets is the fact of having ``ground truth'' labelling available. Indeed, the fact that ratings range from negative to positive allows for classification of nodes as trusted or not trusted. For the two considered graphs, the ``ground truth'' is constructed the following way. After rescaling ratings between $-1$ and $1$, trusted users are the founders of the platform and market players that are rated positively by them (ratings at least $0.5$), whereas fraudulent are these rated negatively (ratings at most $-0.5$) by benign users~\cite{kumar2018rev2}.

\begin{table}
\begin{center}
\begin{tabular}{lrr}
\toprule
                      &  \texttt{Bitcoin OTC} &  \texttt{Bitcoin Alpha} \\
\midrule
               \# of Nodes &     5881 &       3783 \\
              \# of Edges &    35592 &      24186 \\
           Min Degree &        1 &          1 \\
           Max Degree &     1298 &        888 \\
           Avg Degree &       12.10 &         12.79 \\
        Median Degree &        4 &          4 \\
        99th Quantile Degree &      142.20 &        152.36 \\
         \# of Components &        4 &          5 \\
Size of the largest component &     5875 &       3775 \\
           \# of Isolated Nodes &        0 &          0 \\
             Clustering coefficient-global &        0.045 &          0.064 \\
              Clustering coefficient-local &        0.151 &          0.158 \\
\bottomrule
\end{tabular}
\caption{Basic statistics of \texttt{Bitcoin OTC} and \texttt{Bitcoin Alpha} datasets.}
\label{table:graph_stats}
\end{center}
\end{table}

\medskip

The general outline of the experiment presented in this section is as follows. We selected two popular classical embeddings, \textbf{Node2Vec} and \textbf{DeepWalk}, and two structural embeddings, \textbf{RolX} and \textbf{Struct2Vec}. Next, we selected the following widely used six graph features: betweenness, closeness, degree, harmonic centrality, local clustering coefficient, and pagerank. Our objective was to check if these four considered embeddings can be used to predict the values of the selected features. For each pair consisting of an embedding and a selected feature, a predictive model was built. In such predictive models, node feature is a target variable and embedding provides explanatory variables.

In the experiment, graph features were scaled to be in range between 0 and 1 to ensure comparability between them. We have tested multiple dimensionalities of the considered embeddings. Below, we only present the results for 64 dimensional ones, as the results are very similar for all tested dimensions (4, 16, 32, 64, 128). We used two separate datasets (\texttt{Bitcoin OTC} and \texttt{Bitcoin Alpha} from \cite{kumar2018rev2, kumar2016edge}) for the same task to make sure the conclusions are robust.

\medskip

Each predictive model was created using \texttt{H2O AutoML}, which chose the best model in automated manner using the cross-validated mean residual deviance metrics from the set of following models: XGBoost, Deep Learning, and GBM. In the paper we only report test-set $R^2$ for all models for predicting graph features. However, using other metrics such as $RMSE$, $MAE$ leads to similar conclusions.

\begin{table}
\begin{center}
{\footnotesize
    \begin{tabular}{ |p{4cm}|p{1.8cm}|p{1.8cm}|p{1.8cm}|p{1.8cm}|  }
    \hline
    \multicolumn{1}{|c|}{ } &
    \multicolumn{2}{|c|}{Classic} & \multicolumn{2}{|c|}{Structural}\\
    \hline
    Node feature &  Node2Vec &  Deepwalk &  RolX &  Struc2Vec \\
    \hline
    \textit{betweenness}           &      -0.01 &     0.00 &  0.38 &       -0.02 \\ \textit{closeness}             &      -0.02 &     -0.01 &  0.38 &       -0.02 \\ \textit{degree}                &      -0.01 &     -0.01 &  0.59 &       -0.01 \\ \textit{harmonic centrality}   &      -0.01 &      -0.01 &  0.92 &       0.08 \\ \textit{local clustering coef} &      -0.01 &       -0.01 &  0.53 &       0.01 \\ \textit{pagerank}              &      -0.01 &     -0.01 &  0.53 &       0.01 \\
    \hline
    \end{tabular}
    \caption{Embedding's explainability in terms of $R^2$ for \texttt{Bitcoin OTC} dataset.}
    \label{table:model_stats_bitcoin_otc}
    }
\end{center}
\end{table}

\begin{table}
\begin{center}
{\footnotesize
    \begin{tabular}{ |p{4cm}|p{1.8cm}|p{1.8cm}|p{1.8cm}|p{1.8cm}|  }
    \hline
    \multicolumn{1}{|c|}{ } &
    \multicolumn{2}{|c|}{Classic} & \multicolumn{2}{|c|}{Structural}\\
    \hline
    Node feature &  Node2Vec &  Deepwalk &  RolX &  Struc2Vec \\
    \hline
    \textit{betweenness}           &      0.00 &      0.00 &  0.26 &       0.04 \\ \textit{closeness}             &      0.00 &      0.00 &  0.26 &       0.04 \\ \textit{degree}                &      0.00 &      0.00 &  0.51 &       0.11 \\ \textit{harmonic centrality}   &      0.19 &      0.24 &  0.87 &       0.61 \\ \textit{local clustering coef} &      0.01 &      0.00 &  0.48 &       0.09 \\ \textit{pagerank}              &      -0.01 &         0.01 &  0.47 &       0.08 \\
    \hline
    \end{tabular}
    \caption{Embedding's explainability in terms of $R^2$ for \texttt{Bitcoin Alpha} dataset}
    \label{table:model_stats_bitcoin_alpha}
    }
\end{center}
\end{table}

Tables~\ref{table:model_stats_bitcoin_otc} and~\ref{table:model_stats_bitcoin_alpha} illustrate the potential diversity in the results one can obtain in such an experiment. Let us summarize the major conclusions that can be drawn:
\begin{itemize}
    \item in general, classical embeddings are not useful for prediction of graph features (the only exception in our tests is harmonic centrality which for one of the graphs is predictable);
    \item structural embeddings outperform classic embeddings in prediction of graph features;
    \item for a selected structural embedding, there can be significant differences in its performance for different graphs, so it is not possible to draw a general conclusion on exact properties of a given structural embedding a priori;
    \item there can be significant differences in performance of different structural embeddings for the same graph; in our experiment \textbf{RolX} performed much better than \textbf{Struct2Vec}, but for other graphs the relation might be different.
\end{itemize}

For these reasons, if a data scientist has a concrete problem at hand, for which a good structural embedding is needed, it is impossible to decide a priori which such embedding should be chosen. Since there are many possible structural embeddings to choose from (and even for a single embedding algorithm there are usually many parameters that can be tuned), it is desirable to be able to filter-out structural embeddings that are not likely to perform well in a given task at hand (in our example from this section \textbf{Struct2Vec} embedding seems to be not so promising). Therefore, in the next section we propose a framework that allows for creation of a ranking of structural embeddings so that the user can concentrate on analysis of only selected best ones.

\section{Framework}\label{sec:framework_introduction}

In this section we introduce our framework and highlight its properties. We start with providing a high level overview (Subsection~\ref{sec:big_picture}) before formally presenting our algorithm (Subsection~\ref{sec:formal_description_of_the_algorithm}). Justification for some specific design choices and theoretical properties are discussed next (Subsection~\ref{sec:properties}) and we conclude this section with some initial experiments (Subsection~\ref{sec:initial_experiments}).

\subsection{Input/Output}\label{sec:big_picture}

The goal of the framework is to evaluate possible correlations of various node embeddings with a number of classical node features of a single graph $G=(V,E)$ on $n = |V|$ nodes. The input consists of
\begin{itemize}
    \item $k$ dimensional node embedding---$k$ vectors of real numbers, each of length $n$,
    \item $\ell$ node features---$\ell$ vectors of real numbers, each of length $n$.
\end{itemize}
The framework outputs the following 
\begin{itemize}
    \item a real number (represented by symbol $\psi$) from the interval $[0, 1]$ representing how well given feature vectors may be approximated by given embedding vectors; $\psi = 0$ indicates a good approximation and the other extreme value, $\psi=1$, represents a bad approximation; both pre- and post-optimization values of $\psi$ are returned, where the post-optimization $\psi$ value is computed by minimizing $\psi$ as a function of a vector \textbf{w}---formal definition and more details will be provided soon,    
    \item a vector $\textbf{w}$ of non-negative real values of length $k$ and $L^1$-norm equal to $1$ that indicates which embedding dimensions contribute to the explanation of features; here, larger values correspond to larger contribution; the $\textbf{w}$ vector consists of the weights in the embedding distance computation, and is used to identify which embedding dimension the structural feature is mapped onto.
\end{itemize}

The structure of our framework is designed to output a quantitative metric $\psi$, which measures how well an embedding algorithm has learned a given feature (or a collection of features). This metric can be used to both identify what features embedding algorithms learn, in addition to how well they learn those features. A more comprehensive explanation of this is given in the following section.

\subsection{Formal Description of the Algorithm}\label{sec:formal_description_of_the_algorithm}

In our framework, nodes are clustered (using $k$-means clustering) in the feature-space, and distance between sampled nodes in the feature-space are calculated and compared to the distance measured in the embedded-space. Therefore, the algorithm has a few parameters that the user might experiment with but each of them has a default value:
\begin{itemize}
    \item $s$: the number of clusters in the feature space generated by the $k$-means algorithm (by default, $s=\sqrt n$, where $n$ is the number of nodes of a network); the value $s=\sqrt n$ is a safe estimated to ensure the convergence and stability of the calculated $\psi$ metric---more on this is discussed in the following section,
    \item $p$: the fraction of sampled pairs of nodes that are from the same cluster (by default, $p = 0.5$),
    \item $c$: the total number of sampled pairs of nodes (by default, $c = \min\{10^5, n^2/s\}$; apart from a natural upper bound of $10^5$, for small networks we need to make sure that the number of pairs of nodes sampled within clusters, $p\cdot c$, is at most the number of all pairs of nodes from the same cluster; indeed, at the worst case scenario each cluster could consist of $n/s$ nodes and so there could be only $\binom{n/s}{2} \cdot s \approx n^2/(2s)$ pairs of nodes within clusters; this would cause a problem as the algorithm samples pairs without replacement),
    \item standardization method: we provide two methods, MinMax that scales and translates each feature individually such that all of them are in the range between zero and one, and StandardScaler that scales features such that the mean and the standard deviation are equal to zero and, respectively, one (by default, we use the StandardScaler normalization).
\end{itemize}

The algorithm performs the following steps:

\begin{enumerate}
    \item \textbf{Standardization}. Transform all feature and embedding vectors using one of the two methods, MinMax or StandardScaler. 
    After this transformation, all vectors are appropriately normalized and standardized. As a result, later steps are invariant with respect to any affine transformation of these vectors.     
    \item \textbf{Clustering}. Perform the classical $k$-means clustering of nodes (into $s$ clusters) in the feature space using the selected metrics. Let $(c_1, \ldots, c_s)$ with $n = \sum_{i=1}^s c_i$ be the distribution of cluster sizes.
    \item \textbf{Sampling}. There are two types of pairs of nodes that are independently sampled as follows. 
    \begin{itemize}
    \item [a)] sample $$ \hat{m} = \min \left\{ \Big \lfloor p\cdot c \Big\rfloor,\sum_{1 \le a \le s} \binom{c_a}{2} \right\}$$ unique pairs of nodes within clusters; a single pair of nodes is sampled by first selecting cluster $i$ of size $c_i$ with probability equal to $$p(i) = \frac {\binom{c_i}{2}-x_i}{\sum_{1 \le a \le s} (\binom{c_a}{2}-x_a)},$$
    where $x_i$ is the number of pairs already sampled from cluster $i$,
    and then selecting a pair of nodes from the chosen cluster, uniformly at random; if a pair of nodes sampled this way is already present in the sampled set we discard it, otherwise we keep it.
    \item [b)] sample $$ \bar{m} = \min \left\{ \Big\lfloor(1-p)\cdot c \Big\rfloor, \sum_{1 \le a<b \le s} c_a c_b \right\}$$ unique pairs of nodes that are between clusters; a single pair of nodes is sampled by first selecting two clusters $i,j$ ($i < j$) with probability equal to $$p(i,j) = \frac {c_i c_j - x_{i,j}}{\sum_{1 \le a<b \le s} (c_a c_b - x_{a,b})},$$ where $x_{i,j}$ is the number of pairs between cluster $i$ and cluster $j$ already sampled, and then selecting one node from each of the chosen clusters, uniformly at random; if a pair of nodes sampled this way is already present in the sampled set we discard it, otherwise we keep it.
    \end{itemize}        
    \item \textbf{Computing Feature Distance}. For each of the sampled pairs of nodes, compute the corresponding distance in the $\ell$-dimensional feature space $d_f$. For the Euclidean metric we have
    $$ d_f(v_i,v_j) = \sqrt{ \sum_{1 \le a \le \ell}  (f^i_a - f^j_a)^2},$$ 
    where $(f^i_1, \ldots, f^i_\ell)$ and $(f^j_1, \ldots, f^j_\ell)$ are features of nodes $v_i$ and, respectively, $v_j$.
    \item \textbf{Computing Embedded Distance}. Suppose for a moment that a normalized vector of non-negative weights $\textbf{w} = (w_1, \ldots, w_{k})$ with $\sum_{i=1}^{k} w_i = 1$ is fixed. For each of the sampled pairs of nodes, compute the corresponding distance in the $k$-dimensional embedded space $d_e$. The weighted Euclidean distance is given by
    $$ d_e(v_i,v_j) = \sqrt{ \sum_{1 \le a \le k}  w_a (e^i_a - e^j_a)^2},$$ 
    where $(e^i_1, \ldots, e^i_k)$ and $(e^j_1, \ldots, e^j_k)$ are embeddings of nodes $v_i$ and, respectively, $v_j$.
    \item \textbf{Correlation between the two spaces}. To compute the correlation between the two spaces, we define a metric $\psi = 1-r^2 \in [0,1]$, where $r \in [-1,1]$ is the Pearson correlation between vectors in the embedding space and vectors in the feature space. As a result, $\psi$ is defined such that both large positive (close to 1) and large negative (close to -1) correlation would have small values (close to 0). This is done so that the optimization scheme (see the next bullet-point) is more stable.
    \item \textbf{Optimization}. Optimize vector $\textbf{w}$ to minimize $\psi$, where the final value of $\psi$ is referred to as the post-optimization $\psi$. These optimized vectors reflect the importance of embedded dimensions for selected features. We note that the optimization is done using Quasi-Newtonian bounded constraint minimization technique from \textit{Scipy} Optimize method. We note that the pre-optimization $\psi$ value measures the overall raw embedding of a particular feature. To measure how well a feature is learned by a particular embedding algorithm, $\psi$ is optimized against that feature. The optimization process removes (or minimizes) any embedded information that does not contribute to the representation of the feature at study. Therefore, we use the post-optimization $\psi$ value to conduct all experiments in this study.
\end{enumerate}

\subsection{Properties}\label{sec:properties}

Let us briefly highlight some basic and desired properties of the framework which, in particular, justify its design and show its potential usefulness. 

\begin{itemize}
    \item The framework is designed in such a way that affine transformations of any of the feature or embedding vectors do not change the results.
    \item The framework does not assume any particular type of the relationship between feature space and embedded space. Instead, it is desired that if two nodes are close in the feature space, then they are also close in the embedded space (with a proper metrics/weighting in the embedded space).
    \item The sampling strategy used in the framework has the following consequence. Achieving a good $\psi$ score ensures that close pairs of nodes in the feature space are close in the embedded space. On the other hand, if some pairs of points are far in the feature space, then the framework puts less weight on the fact whether they are close or not in the embedded space. The rationale behind this property is that a typical pair of nodes are likely to be far in both spaces and so the framework should not pay too much attention to these pairs. 
    \item An embedding algorithm might learn many node features, which may not contribute to the representation of particular structural feature. This additional learned information can be removed and minimized by adjusting the weights associated with appropriate embedding dimensions. For example, the feature \textit{PageRank} may get mapped to dimension of 1 (out of 8) of an embedding space. In this case, dimensions 2 to 7 do not contribute to the representation of \textit{PageRank} and can be removed by setting the weights for those dimensions to 0. This process is done automatically by our framework during the optimization process of the $\psi$ value.
    \item It is the responsibility of the analyst to select features that should be represented well in the correspondence between the feature space and the embedding space. The assumption here is that each feature is equally important. However, the analyst may, on top of passing all features to the framework, pass each of them independently, one by one, to assess which embedding dimensions are important for a given feature.
    \item All steps performed by the framework are standard, well-known, and easily available in most data science frameworks. As a result, users might easily modify the framework, if needed, to incorporate any specific needs or situations that are encountered. Moreover, since all of these steps are standards, they are typically implemented efficiently which makes the framework to be relatively fast.
\end{itemize}

\section{Experimentation}

\medskip

In this section, we investigate and analyze various desired algorithmic properties of our framework. We focus on six embedding algorithms, four structural ones (\textbf{LSME}~\cite{LSME}, \textbf{Role2Vec}~\cite{role2vec}, \textbf{Struc2Vec}~\cite{struc2vec}, and \textbf{RolX}~\cite{rolx}) and two classical ones (\textbf{Node2Vec}~\cite{node2vec} and \textbf{DeepWalk}~\cite{deepwalk}). The goal of our analysis here is to understand and analyze various properties of our framework. In addition, we showcase how one could use our framework in applications such as node classification, by investigating the performance of a number of node and structural embedding algorithms. We break up our analysis into two main parts. First (Subsection~\ref{sec:initial_experiments}), we explore some of the basic properties of our framework, such as algorithm stability and behaviour. Second (Subsection~\ref{sec:role_classification_case_study}), we showcase the application of our framework in a node classification case study. In this section, we use the default hyper-parameters for every embedding algorithm. Lastly, we use both a synthetically designed graph in addition to real-world networks for the role classification task.

\subsection{Synthetic Network}\label{sec:graphs_and_datasets}
\begin{figure}[h]
  \centering
  \includegraphics[scale=0.33]{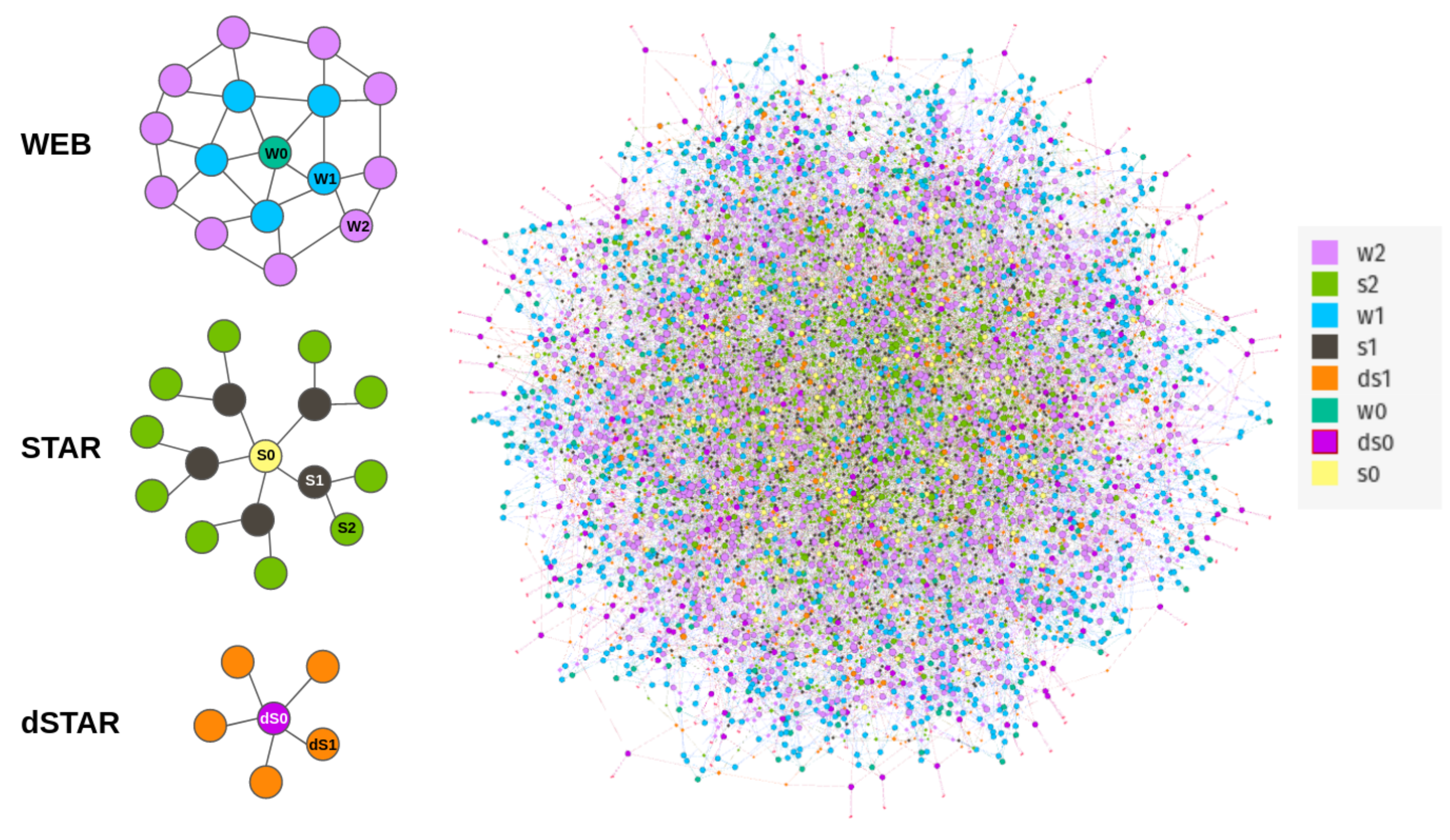}
  \caption{Synthetic graph $\mathcal{G}(\{N_w=200, N_s=200, N_{ds}=200\}, \{k_{w1}=5, k_{w2}=10\}, \{k_{s1}=5, k_{s2}=10\}, \{k_{ds1}=5\})$ composed of a collection of Web, Star and dStar subgraphs.}
  \label{fig:dense_star_graph_network_with_example}
\end{figure}

For experiments in this section we use synthetically generated graph $\mathcal{G}$ which is composed of three structurally distinct sets of subgraphs. As shown in Figure \ref{fig:dense_star_graph_network_with_example}, these subgraphs are labelled Web, Star and dStar. The Web and Star subgraphs each have three types of nodes (w0, w1 and w2) and (s0, s1, s2), while dStar subgraph has two types of nodes (ds0, ds1). The overall synthetic graph is created by joining $N_w$ Web, $N_s$ Star and $N_{ds}$ dStar subgraphs by randomly creating links between w2, s2 and ds1 nodes. The edge creation process is as follows; from joined set of w2, s2 and ds1 nodes randomly select two nodes $n_a$ and $n_b$. If $n_a \neq n_b$ and $e_{ab} \notin E$, where $E$ is the set of edges of $\mathcal{G}$, then create an edge $e_{ab}$. Repeat this process until all w2, s2 and ds1 nodes are connected to at least one other node. Based on this description, we can fully define our synthetic graph using 8 parameters: $\mathcal{G}(\{N_w, N_s, N_{ds}\}, \{k_{w1}, k_{w2}\}, \{k_{s1}, k_{s2}\}, \{k_{ds1}\})$. Here, $N_w, N_s, N_{ds}$ are the number of Web, Star and dStar subgraphs in the overall graph and, $k_x$ correspond to the number of nodes in layer $x$ of each subgraph. Each layer of the subgraphs is connected to the previous/next layers as shown in Figure \ref{fig:dense_star_graph_network_with_example}. For example, $k_{w1} = 5$, based on Figure \ref{fig:dense_star_graph_network_with_example}. In this section, we create a synthetic graph with the following parameters: $\mathcal{G}(\{N_w=200, N_s=200, N_{ds}=200\}, \{k_{w1}=5, k_{w2}=10\}, \{k_{s1}=5, k_{s2}=10\}, \{k_{ds1}=5\})$, resulting a synthetic graph with 7,600 nodes. We have chosen this structure for our synthetic graph to allow for a simple yet structurally distinct nodes to be used for our classification tasks. Since our framework is designed for structural embedding algorithms, we wanted to use synthetic graphs where nodes have known structural roles (ground-truth). As we shall show in Section~\ref{sec:role_classification_case_study}, we use the synthetic graph described above to build classifiers for identifying root nodes $S_0$, and analyze each embedding algorithm's performance for this task.

\subsection{Real-World Network}\label{sec:graphs_and_datasets_twitch}
In addition to the synthetic network described above, we will use a real-world network constructed using Twitch social network. Twitch is a social network app/website on which users primarily share video-game streams and commentary~\cite{rozemberczki2021twitch}. Here, we use a sample of this network, where nodes correspond to users and the edge between two nodes indicates that the users follow one another. In addition to the graph data, there are meta-data associated with each user. A breakdown of the graph statistics and user meta-data is presented in tables \ref{table:twitch_graph_statistics} and \ref{table:twitch_metadata_description}, respectively.

\begin{table}
\begin{center}
\begin{tabular}{lrr}
\toprule
\textbf{Metric} &  \textbf{Stat} \\
\midrule
Number of Nodes & 168,114\\
Number of Edges & 6,797,557\\
Min Degree & 1 \\
Max Degree & 35,279\\
Avg Degree & 80.86\\
Number of Components & 1\\
\bottomrule
\end{tabular}
\caption{Statistical properties of the Twitch network.}
\label{table:twitch_graph_statistics}
\end{center}
\end{table}

For the purpose of our analysis, we will use \textit{mature} and \textit{views} as targets for the predictive case study. The \textit{mature} status of the accounts will be used for a classification task, where we will use node features, in addition to the embedding feature to classify whether a user's account is labeled mature or not. Similarly, for the \textit{views} count, we will build regression models to predict the number of views an account gets, based on features derived from node and embedding features. In addition to building and analyzing classification and regression models, we use our framework to extract insight into these tasks. The results for this work is presented in section \ref{sec:application_of_framework_to_real_network}. 

\begin{table}
\begin{center}
\begin{tabular}{lrr}
\toprule
\textbf{Meta-Data Feature} &  \textbf{Description} & \textbf{Type} \\
\midrule
views & View count estimation & Numerical\\
life time & User lifetime estimation & Numerical\\
mature & Explicit content streamer identification& Categorical\\
dead account & Churn prediction & Categorical\\
affiliate & Affiliate status identification & Categorical\\
language & Language of the account & Categorical\\
created at & Account creation date & Date\\
update at & Latest update to the account & Date\\
\bottomrule
\end{tabular}
\caption{Meta-data features for the Twitch dataset~\cite{rozemberczki2021twitch}.}
\label{table:twitch_metadata_description}
\end{center}
\end{table}

\begin{figure}[h]
  \centering
  \includegraphics[scale=0.55]{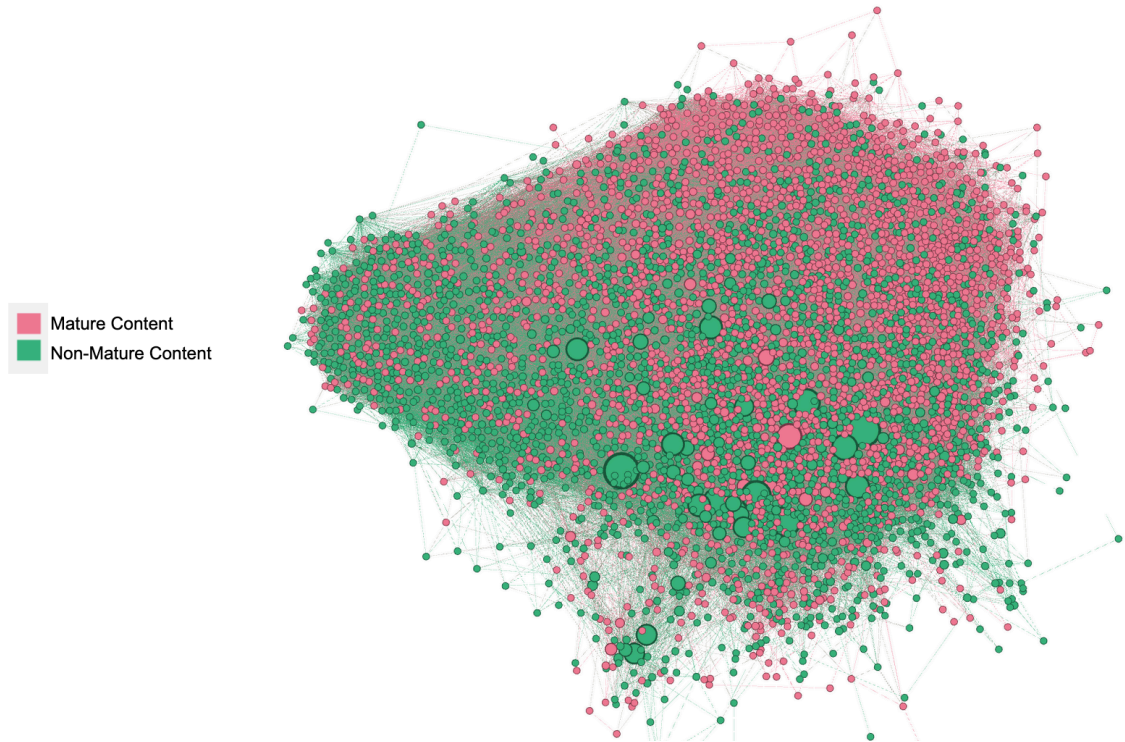}
  \caption{Network structure for a sample of 5,000 nodes from the Twitch network. Red nodes correspond to users with mature content and green nodes correspond to users with non-mature content. The size of each node is proportional to the number of views each user has on Twitch.}
  \label{fig:twitch_graph}
\end{figure}

\subsection{Algorithmic Properties of the Framework}\label{sec:initial_experiments}

In this section, we analyze various algorithmic properties of our framework such as the convergence and stability of various metrics and the behaviour of structural vs.\ classical embedding algorithms. As we described in Section~\ref{sec:formal_description_of_the_algorithm}, the quality of learned representation of a structural feature (for example degree centrality) is measured using the post-optimization $\psi$ value. The optimization is done by minimizing $\psi$ as a function of weights associated with each embedding dimension. To test the effectiveness of this approach, we performed two experiments. In each experiment, we embed the synthetically created graph described in the previous section using either \textbf{LSME} or a fixed-embedding algorithm. Here, the fixed-embedding maps the simplest centrality measure, degree centrality, directly onto one of the $N$ embedding dimensions. Furthermore, the other $N-1$ dimensions are filled with random numbers. This simulates a synthetic embedding algorithm, which learns a perfect representation of a feature and maps it onto one of the dimensions of the embedding space. For our experiments, we used $N=8$ as the dimension of both embedding algorithms. Once the embedding vectors are generated, we use our framework to measure the performance of each embedding with respect to the degree centrality. In other words, we measure how well did each embedding learn the representation of degree centrality. The purpose for this experiment is to showcase the optimization process and highlight how our framework can be used to study how well features are mapped into the embedded space.

\begin{figure}[h]
  \centering
  \includegraphics[scale=0.7]{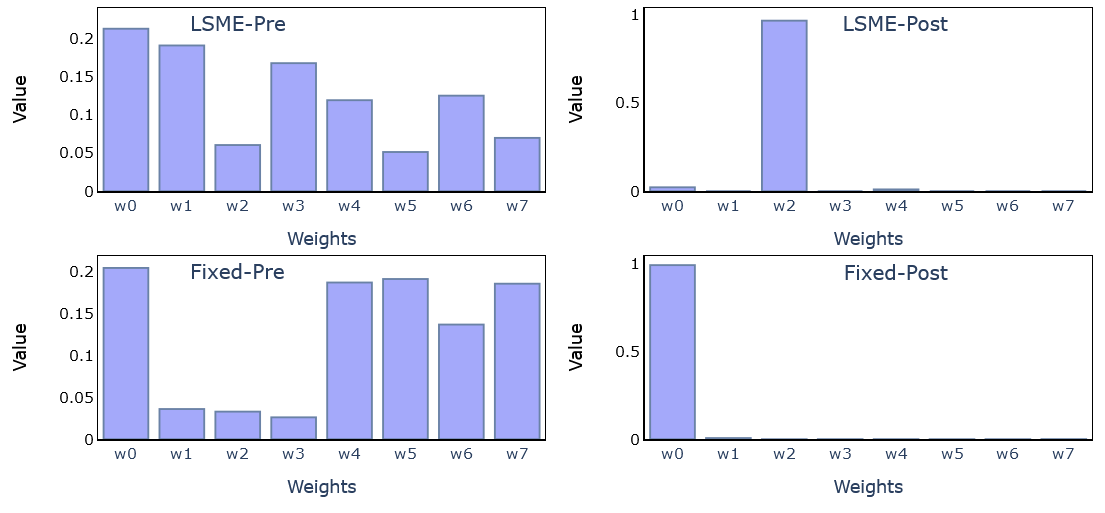}
  \caption{Pre and post optimization values for weights associated with each embedding dimension (8 dimensional embedding). Top-left and bottom-left figures show pre-optimization random initialization of the weights for \textbf{LSME} embedding and fixed embedding, respectively. Top-right and bottom-right are the post-optimization values of the weight for \textbf{LSME} and fixed embedding.}
  \label{fig:pre_and_post_optimization_weights}
\end{figure}

Figure~\ref{fig:pre_and_post_optimization_weights} shows the results for the pre and post optimization values for the weights associated with each embedding dimension. In the pre-optimization state (top-left and bottom-left), the values for the weights are set randomly. The optimization algorithm then identifies the dimensions which the representation of the degree centrality was mapped onto. As expected, the post-optimization weights for the fixed embedding is collapsed to only dimension 0 (w0), which holds a copy of degree centrality for the nodes. For the \textbf{LSME} algorithm, degree centrality was mapped by the embedding algorithm onto primarily dimension 2 and partially onto dimension 0 and 4. It is important to note that the post optimization weights by themselves are not complete measure of how well the embedding has learned the node feature. To capture the complete picture, one has to also consider the post optimization score $\psi$---more on this soon. The experiments in Figure \ref{fig:pre_and_post_optimization_weights} were repeated multiple times with randomized initial weights, and the results of the experiments were consistent with the above findings.

\medskip

\begin{figure}[h]
  \centering
  \includegraphics[scale=0.6]{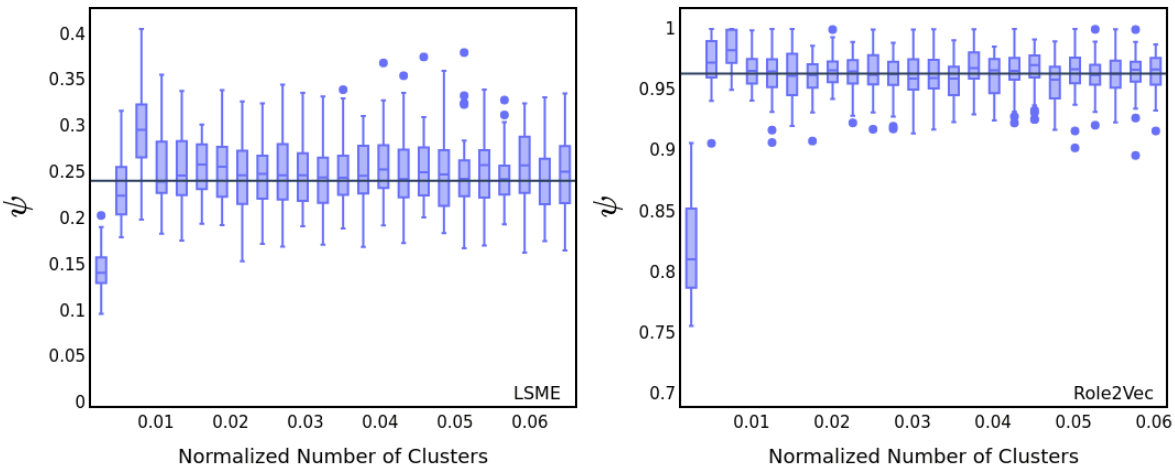}
  \caption{Post-optimization score $\psi$ as a function of normalized number of clusters. Normalization is with respect to the size of the graph. On the left/right $\psi$ is computed for the degree centrality using the \textbf{LSME}/\textbf{Role2Vec} embedding algorithms, respectively. The horizontal lines are the long-run average of $\psi$.}
  \label{fig:score_stability_as_a_function_of_number_of_clusters}
\end{figure}

\begin{figure}[h]
  \centering
  \includegraphics[scale=0.6]{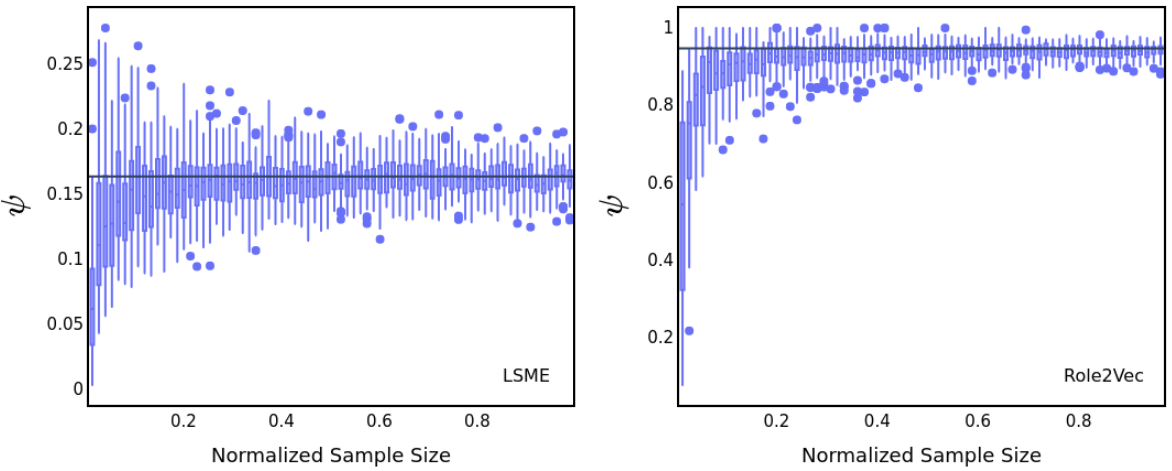}
  \caption{Post-optimization score $\psi$ as a function of normalized sample size. Normalization is with respect to the size of the graph. On the left/right $\psi$ is computed for the degree centrality using the \textbf{LSME}/\textbf{Role2Vec} embedding algorithms, respectively. The horizontal lines are the long-run average of $\psi$.}
  \label{fig:score_stability_as_a_function_of_sample_size}
\end{figure}

Before we dive deeper into the other properties of the framework, we want to consider the stability of the algorithm as a function of the node sample size (parameter $c$) and the number of clusters produced by $k$-means algorithm (parameter $s$). To apply the framework to large graphs, we would want our algorithm to converge for both $c \ll n$ and $s \ll n$, where $n$ is the number of nodes of the graph. To measure the stability of the algorithm, we perform two experiments using \textbf{LSME} and \textbf{Role2Vec} on a synthetic graph. The performance of the embedding algorithms are measured using the degree centrality as the node feature. Figures~\ref{fig:score_stability_as_a_function_of_number_of_clusters} and~\ref{fig:score_stability_as_a_function_of_sample_size} show the convergence of $\psi$ as a function of the normalized number of clusters and the normalized sample size respectively, where normalization is done with respect to the size of the graph. Let us first consider the behaviour of $\psi$ as a function of $s$, the number of clusters. It is important to note that while we vary $s$ in this experiment, values for other parameters are kept as default. As we can see in Figure~\ref{fig:score_stability_as_a_function_of_number_of_clusters}, $\psi$ converges to its long-run average when the number of clusters is approximately 2\% to 3\% of the total size of the graph. Both here and in Figure~\ref{fig:score_stability_as_a_function_of_sample_size}, the long-run average is defined as the expected value for $\psi$ as sample size or number of clusters approaches the size of the graph. Finally, we conclude that our default value for the number of clusters ($s = \sqrt{n}$) is a good approximation since for the current experiment ($n=1,000$) the number of clusters is approximately 3\% of the size of the graph ($\frac{\sqrt{1,000}}{1,000} \approx 0.03$). Next, we look at the convergence of $\psi$ as a function of $c$, the sample size. Similarly to the previous experiment, we vary $c$ while setting other parameters to their default values. As one can see in Figure~\ref{fig:score_stability_as_a_function_of_sample_size}, the value for $\psi$ converges for sample of sizes greater than or equal to 20\% to 30\% of the size of the graph. The results of our experiments point at two facts. First that the algorithm converges and is stable for both $c \ll n$ and $s \ll n$. Second, the default values for the hyperparameters are good and safe estimates.

\medskip

We now turn our attention to experiments comparing the general behaviour of structural embedding as compared to classical node embedding algorithms. As we highlighted in Section~\ref{sec:motivation}, classical node embedding algorithms such as \textbf{Node2Vec}~\cite{node2vec} and \textbf{DeepWalk}~\cite{deepwalk} have difficulty learning structural properties of graphs. To showcase that our framework can be used as an unsupervised method for capturing this fact, we perform six experiments using four structural and two classical embedding algorithms. All six algorithms are ran against synthetic graphs (created using the procedure in Section \ref{sec:graphs_and_datasets}) to generate 8-dimensional embedding vectors. Each embedding is then evaluated using our framework, where its performance is measured against 12 classical and widely used node features. For each node feature, we compute the post optimization $\psi$ value. As we noted previously, the value of $\psi$ is inversely correlated to how well the embedding was able to learn a given representation. In particular, $\psi = 1$ means that the embedding was not able to learn anything for a given feature, and in the other extreme $\psi = 0$ means that the embedding was able to learn prefect representation of the feature. With that said, let us consider the results presented in Figure~\ref{fig:score_dense_star}. Here, we show the post-optimization value for $\psi$ as a function of various node features for \textbf{LSME}, \textbf{Struc2vec}, \textbf{RolX} and \textbf{Role2Vec}, respectively. It is clear that these structural embedding algorithms performed differently, as measure by our framework. The \textbf{LSME}, \textbf{Struc2Vec} and \textbf{RolX} algorithms performs much better than \textbf{Role2Vec}. It is important to note that we have chosen to use the default settings and parameters for each algorithm and did not perform any optimization. In addition, we have only focused on a set of 12 structural features, while algorithms could be learning features not in our set. While the structural embedding algorithms \textbf{LSME}, \textbf{Struc2Vec} and \textbf{RolX} were able to learn some structural node features, as expected, classical embedding algorithms (\textbf{Node2Vec} and \textbf{DeepWalk}) struggle with this task. This is clearly shown in the top right and top left plots of Figure~\ref{fig:score_dense_star}, where the post-optimization $\psi$ values for all node features are close to 1. Lastly, both \textbf{Node2Vec} and \textbf{DeepWalk} perform similarly to one another, indicating the similarity in the underlying algorithms. It is also important to note that, the performance of these algorithm also depends on the structure of the underlying network. In other words, for an embedding algorithm to learn something about the graph, there must be something to be learned. In our idealized case of a synthetic graph, we have artificially introduced nodes with structural properties to ease the task of learning structural properties for the selected embedding algorithms.

\subsection{Role Classification Case Study}\label{sec:role_classification_case_study}

In this subsection, we explore the use case of our framework for analyzing role classification in a synthetic network introduced earlier. A common task in network analysis is to classify nodes based on the role the nodes play in their local network structure. To build features for a classification algorithm, one could either use manually calculated structural properties of the nodes (node features) or leverage structural or node embedding features; as an automated way of learning various structural properties of nodes. One major challenge with using some embedding algorithms as a source for feature engineering, is the lack of explainability of the learned representations. It is not easy to identify what structural properties of the nodes are learned and how a given learned representation is mapped onto the embedding space. To explore these ideas and showcase one possible use case of our framework, we consider the synthetic network introduced in Subsection~\ref{sec:graphs_and_datasets}, and use w0, s0 and ds0 root nodes as the target nodes we would like to classify. The root nodes considered here have very similar local structure, creating a relatively challenging tasks for a classifier. The goal of our analysis is to design and build a classifier using both node features and features extracted from various embedding algorithms. Lastly, we show how one could use our framework as an unsupervised technique to gain insight into the performance of embedding algorithms in applications such as role classification. We use six embedding algorithms, two classical algorithms (\textbf{Node2Vec} and \textbf{DeepWalk}) and four structural based ones (\textbf{LSME}, \textbf{Struc2Vec}, \textbf{RolX}, and \textbf{Role2Vec}). We hope to answer the following questions: can one use our framework to identify embedding algorithms that best learn various structural properties of nodes, which could hint at their performance in a role classification task? Additionally, can one extract insights into the predictability strength of each node feature and how those features are learned by a given embedding algorithm?   

\begin{figure}[h]
  \centering
  \includegraphics[scale=0.47]{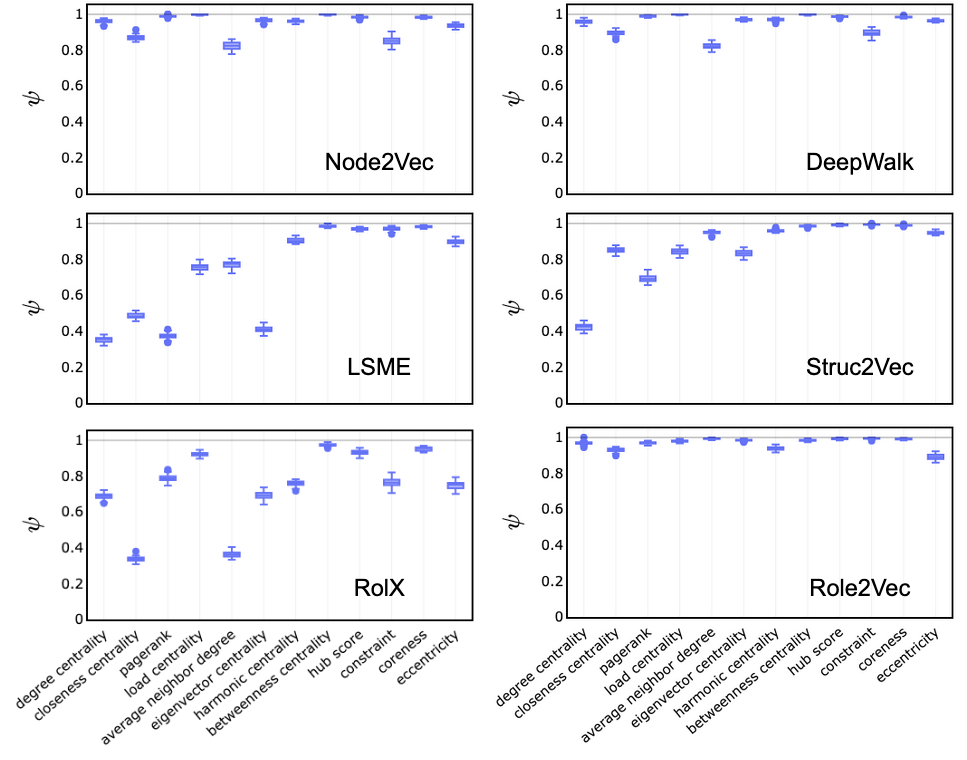}
  \caption{Post-optimization $\psi$ values ($y$-axis) computed as a function of 12 node features ($x$-axis) for various classical and structural embedding algorithms.}
  \label{fig:score_dense_star}
\end{figure}

\medskip

\begin{figure}[h]
  \centering
  \includegraphics[scale=0.5]{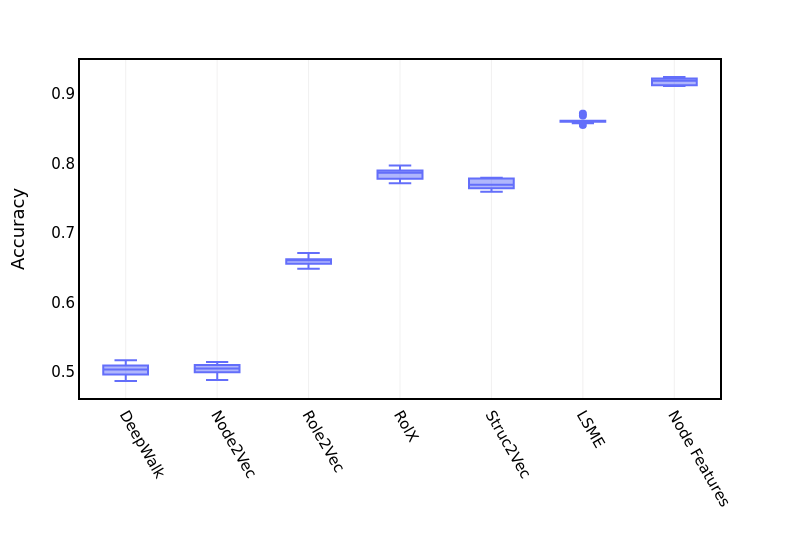}
  \caption{Accuracy of 7 classifiers built using node and embedding features. Here, accuracy is measured as the combined accuracy of the following classes (w0, s0 and ds0).}
  \label{fig:dense_star_model_accuracy}
\end{figure}

We first start by analyzing each embedding algorithm using our framework. We consider 12 node features as a benchmark and compute $\psi$ for each feature. The performance of each embedding algorithm is presented in Figure~\ref{fig:score_dense_star}. As before, $\psi$ is inversely proportional to how well an embedding algorithm has learned a given feature, where $\psi = 1$ means that a given feature was not learned by the algorithm. As we can see in Figure~\ref{fig:score_dense_star}, classical embedding algorithms (\textbf{Node2Vec} and \textbf{DeepWalk}) fail to learn the structural properties of the graphs. This aligns with our expectations, since classical algorithms are designed to learn classical node properties. Furthermore, \textbf{Role2Vec} also fails to learn any structural proprieties of the graph, while \textbf{LSME}, \textbf{Struc2Vec}, and \textbf{RolX} perform quite well. It is important to note that we used the default hyper-parameters for each algorithm, and it is possible to achieve better results if one optimizes the learning process. Using current results, one would expect \textbf{LSME}, \textbf{Struc2Vec}, and \textbf{RolX} algorithms to perform better than \textbf{Node2Vec}, \textbf{DeepWalk}, and \textbf{Role2Vec} in classification tasks, where the structural properties of the nodes are of importance. With the results from Figure~\ref{fig:score_dense_star} in mind, we build 7 classifiers with the goal of classifying w0, s0 and ds0 nodes in graph $G$ using features built using manually computed node features and features from each of our six embedding algorithms. The classifiers are trained to predict 3 classes, one for each root node (w0, s0 and ds0). Note that we do not mix features between embedding algorithms and node-features, in this analysis. For example, the accuracy of the classifier built using \textbf{DeepWalk}, in Figure \ref{fig:dense_star_model_accuracy}, only includes features extracted from the embedding of the nodes by the \textbf{DeepWalk} algorithm.

The overall accuracy of each classifier is plotted in Figure~\ref{fig:dense_star_model_accuracy}. Here, accuracy is measured as the combined accuracy of the following classes (w0, s0 and ds0). For each embedding, we train 10 models and select the best performing model and overage the performance across 10 samples. As expected, based on our analysis in Figure~\ref{fig:score_dense_star}, classifiers built using node features, \textbf{Struc2Vec}, \textbf{RolX}, and \textbf{LSME} perform much better than those built using \textbf{Role2Vc}, \textbf{DeepWalk}, and \textbf{Node2Vec}. It is important to consider the following when analyzing these results. The fact that a classifier built using solely node feature performs well, indicates that any embedding that learns structural properties of the node should also perform well. However, this logic does not apply in reverse. The poor performance of an embedding based on our framework does not necessarily indicate that it will perform poorly in a classification task, since there may be features with predictive power which are not captured by the reference features of our framework. We note that, the set of reference features is modifiable and could be updated to include additional structural features. One could extend the 12 features in the benchmarking set to capture high order structural properties of the graph, to allow for a more extensive list of structural properties. Lastly, we point out that one could use the output of the framework to study the specific features learned by each embedding algorithm. For example, both \textbf{LSME} and \textbf{Struc2Vec} fail to learn \textit{Constrain}, which is the measure of Burt's Constraint \cite{everett2020unpacking} for each node, (see Figure~\ref{fig:score_dense_star}) as a structural feature, while \textbf{RolX} performs better in this regard. This is an important observation in scenarios where one would want to combine the features from different embedding algorithm to built feature sets with more predictive power. It is natural that each embedding algorithm learns slightly different properties of the graph. Our framework can be used as a tool to map out the embedding space and understand it through the lens of structural features of the graph.

\subsection{Application of framework in real-networks}\label{sec:application_of_framework_to_real_network}

In this section, we use our framework to analyze the Twitch dataset (Figure \ref{fig:twitch_graph}). Similar to the synthetic network analysis, we will use the node features and the embeddings to build classification models to identify a particular labeled group of nodes. Here, we will use the "mature" status of the Twitch accounts as the target for the classification task. Note that we do not use node (user) metadata in our model development, as the goal is to study the predictive power of features one could extract using embedding techniques from the underlying social network. The Twitch network has approximately equal number of mature vs non-mature accounts. Similar to the synthetic network, we split our dataset into train, validation and test using the 60/10/30 split fractions, respectively. For each feature source, we train 10 models and the final accuracy is plotted as the distribution of the accuracy of each model.

\medskip

\begin{figure}[h]
  \centering
  \includegraphics[scale=0.5]{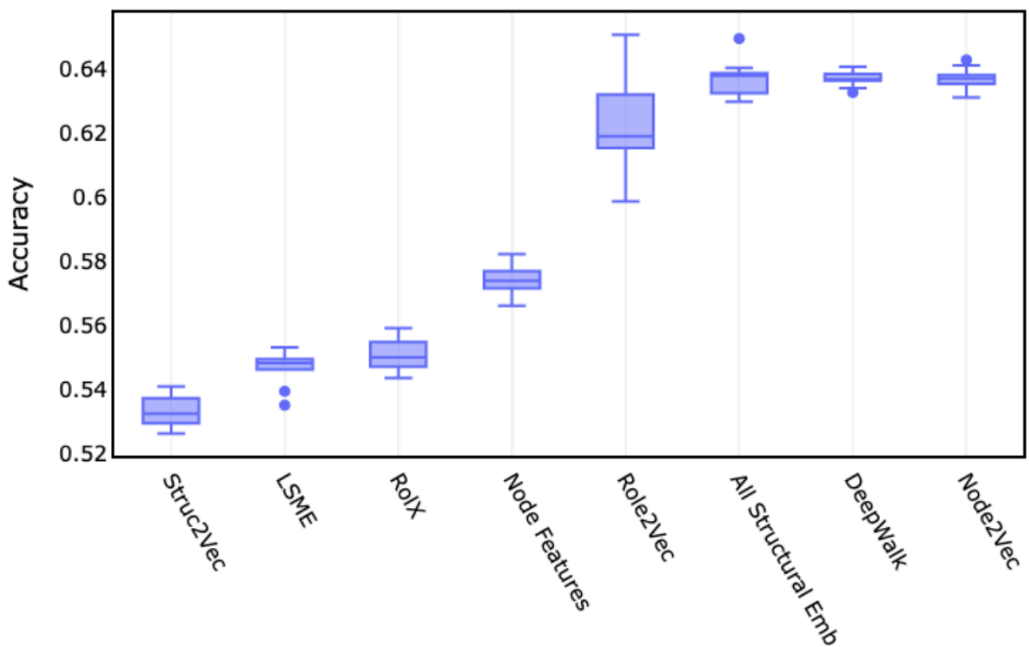}
  \caption{The accuracy of classifiers built to identify Twitch accounts that are labeled "mature". 4 structural (\textbf{Struc2Vec}, \textbf{LSME}, \textbf{RolX}, \textbf{Role2Vec}), 2 classical embeddings (\textbf{DeepWalk}, \textbf{Node2Vec}) are used. In addition we used the node structural features in addition to the combination of all structural features.}
  \label{fig:twitch_mature_classification_accuracy}
\end{figure}

Figure \ref{fig:twitch_mature_classification_accuracy} shows the performance of eight models trained on four structural embeddings (\textbf{Struc2Vec}, \textbf{LSME}, \textbf{RolX}, \textbf{Role2Vec}), two classical embeddings (\textbf{DeepWalk}, \textbf{Node2Vec}), in addition to classifiers trained on node structural features only and lastly a model train on the combination of all structural embeddings. As we can see in Figure \ref{fig:twitch_mature_classification_accuracy}, the classical embedding algorithms (\textbf{DeepWalk}, \textbf{Node2Vec}) perform the best with an accuracy of approximately 64\%. This can be somewhat expected by looking at Figure \ref{fig:twitch_graph}, where the mature (red) and non-mature (green) accounts are roughly divided into two communities. The network layout in \ref{fig:twitch_graph} is arrived at using the \textit{Force Atlas 2} algorithm in the Gephi graph visualization software~\cite{ICWSM09154}. The structural embedding algorithms in general perform worst than the classical embeddings in this particular study. \textbf{Role2Vec} is the only structural embedding algorithm with performance similar to the classical embeddings. The \textbf{Role2Vec} algorithm uses a random walk technique to sample the underlying network, similar to the \textbf{Node2Vec} and \textbf{DeepWalk} embeddings. This could be the reason the \textbf{Role2Vec} algorithm is learning features about the network that other structural embedding algorithms don't. This idea can be further justified as the classifiers built using \textbf{Role2Vec} embedding, outperform classifiers built using node features alone. This hints at the fact that there are either structural properties learned by \textbf{RoleVec} algorithm, which are not present in our 12 node structural features, or that there are non-structural properties of the graph that \textbf{Role2Vec} algorithm learns. 

\begin{figure}[h]
  \centering
  \includegraphics[scale=0.4]{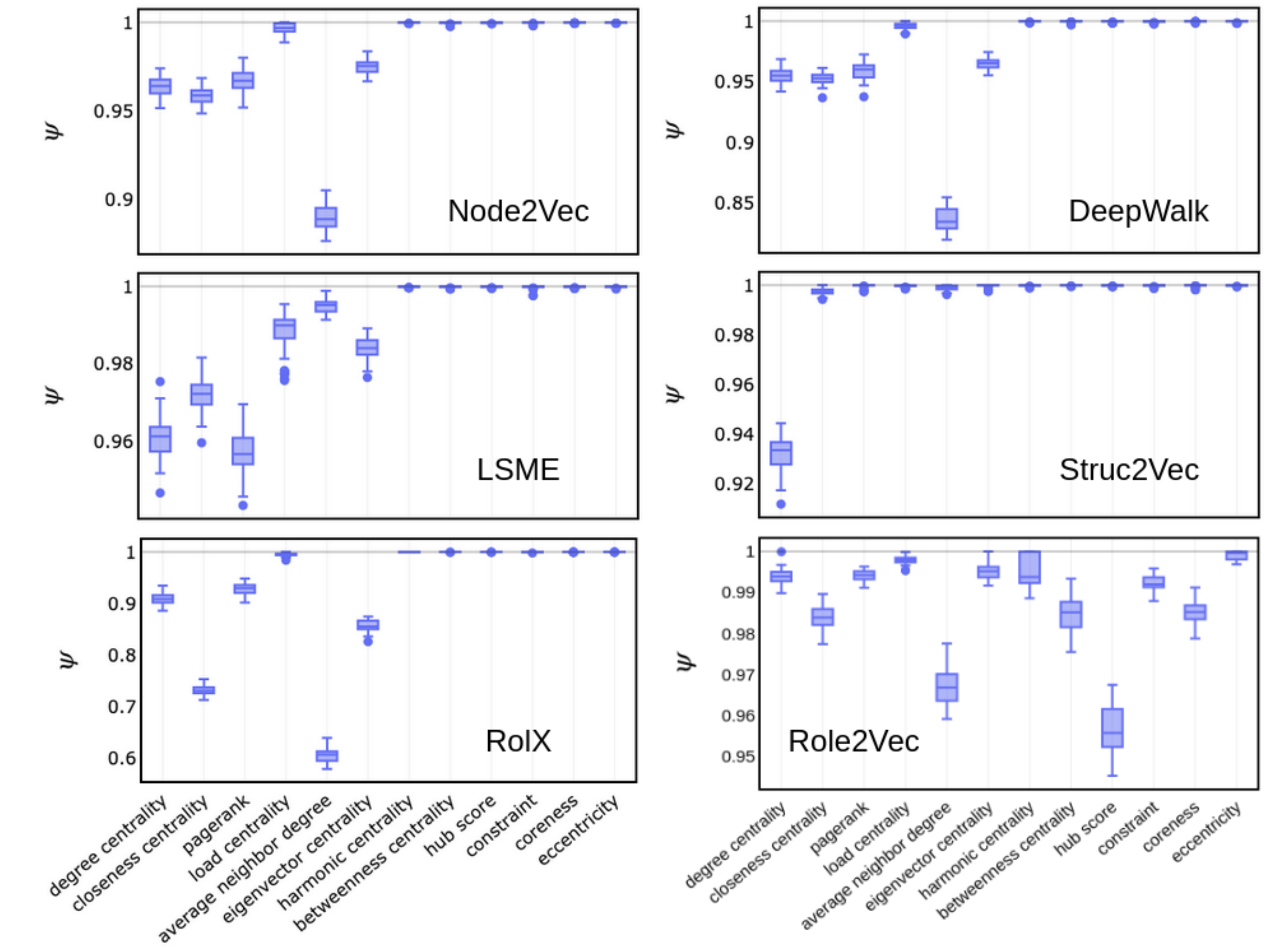}
  \caption{The performance of structural and embedding algorithms measured using $\psi$ for the 12 node structural features.}
  \label{fig:twitch_embedding_psi_values}
\end{figure}

Let us now analyze the behaviour of the embeddings using our framework. We show the performance scores ($\psi$) for the Twitch network in Figure \ref{fig:twitch_embedding_psi_values}. First, we observe that \textbf{Struc2Vec} fails to learn most structural node features, except for \textit{degree centrality}. This aligns with the fact that \textbf{Struc2Vec} performs the worst in the classification task. Secondly, we point out that both \textbf{LSME} and \textbf{RolX} capture a number of structural properties of the network, with \textbf{RolX} performing the best (as measured using $\psi$). This translate to their better perfrormance in the classification task. Lastly, we observe that \textbf{Role2Vec's} performance behaviour, as shown in Figure \ref{fig:twitch_embedding_psi_values}, is the most comprehensive. We can see that \textbf{Role2Vec} learns (at least partially) all 12 node structural features, which translates well to its performance in the classification task. We note however that \textbf{Role2Vec's} performance in the classification task is a result of a combination of learned node structural features in addition to other predictive features not captured by our framework's 12 feature set. This is evident, since \textbf{Role2Vec} outperforms models trained strictly on node structural features (labeled \textit{Node Features} in Figure \ref{fig:twitch_embedding_psi_values}). 

We conclude this section by highlighting the fact that our framework provides useful insights into the behaviour of structural embedding algorithms in capturing structural properties of real networks. As we have shown in Figure~\ref{fig:twitch_embedding_psi_values}, it is easy to rank the structural embeddings based on how they learn various structural properties of the graph, with \textbf{Role2Vec} performing better than \textbf{LSME} and \textbf{RolX}, which in turn perform better than \textbf{Struc2Vec}. Arriving at this information, in an unsupervised way, is useful for practitioners and as we have shown translates well to supervised tasks such as node classification.


\section{Conclusion}
In this work, we introduced an unsupervised embedding evaluation framework which can be used to both explain what structural properties of nodes embedding algorithms learn, in addition to how well each algorithm learns a particular structural feature. As we highlighted in Section~\ref{sec:motivation}, classical embedding algorithms often fail to learn meaningful representation of structural properties. Therefore, for tasks such as role discovery or role classification, we need to rely on structural properties of nodes learned by structural embedding algorithms. As we highlighted, there are numerous challenges with using structural embedding algorithms. First, there is a diverse set of structural features that an algorithm could learn. Therefore, it is not easy to define a single metric for measuring the performance of structural embedding algorithms. Second, measuring performance of embedding algorithms is often done using supervised techniques, which relies on the availability of labeled dataset. In Section~\ref{sec:framework_introduction}, we introduced a framework, which addresses the above two challenges. In our framework, we introduce a collection of core structural features, against which one could measure the performance of a structural embedding algorithm. In addition, we introduce a technique for performing these measurements in an unsupervised way, which avoids the need for the availability of labelled datasets. By introducing a mapping between the embedding and the feature space, we are able to define a metric ($\psi$) for measuring the performance of embedding algorithms. In addition, we can use this metric to explain which features are learned by a given algorithm. As we have shown in Section~\ref{sec:initial_experiments}, this feature of our framework is especially useful for the explainability of algorithms that rely on deep-learning such as \textbf{LSME}. Furthermore, using a synthetic and a real-life (Twitch network) graph as a benchmark, we showcased several use cases for our framework. In Section~\ref{sec:role_classification_case_study}, we showed that one could use our framework to measure the performance of a number of classical and structural embedding algorithms against a set of structural features. The performance of the embedding algorithms, as measured by $\psi$, correlates with the performance of the algorithms in a role classification task. This highlights the utility of our framework, which can be used to gain insights into the performance of embedding algorithms in scenarios where labeled data is not available. In addition, one could use our framework to identify difficult to embed structural features and use the $\psi$ value as iterative way of modifying an embedding algorithm to learn specific feature-sets. The unsupervised framework developed and showcased in this work can be used a versatile tool for practitioners studying structural properties of complex networks.

\bibliographystyle{plain}
\bibliography{bibliography}

\end{document}